# Portraits of Julius Caesar: a proposal for 3D analysis

**Amelia Carolina Sparavigna**
Department of Applied Science and Technology
Politecnico di Torino, C.so Duca degli Abruzzi 24, Torino, Italy

*Here I suggest the use of a 3D scanning and rendering to create some virtual copies of ancient artifacts to study and compare them. In particular, this approach could be interesting for some roman marble busts, two of which are portraits of Julius Caesar, and the third is a realistic portrait of a man recently found at Arles, France. The comparison of some images indicates that a three-dimensional visualization is necessary.*

Until June 25 of this year, the ancient history of Arles is the subject of an exhibition at the Louvre Museum of Paris. The Museum is hosting several Roman pieces recovered from the river Rhone. One of these roman artifacts is the Arles portrait bust [2], a life-sized marble bust, which is the portrait of a man with wrinkles and several other realistic details. Some divers discovered this bust during an archaeological survey in the waters of the river of Arles. It seems that the realism of this portrait is in agreement with the features of plastic arts of the late Republican period, that is, of the first century BCE. This bust was therefore proposed as a portrait of Julius Caesar [2].
If it is dated about 46 BCE, this could be the oldest known representation of Caesar [3]. However, Mary Beard, professor of Classics, remarked the need of further researches for identifying the man as Caesar [4]. In fact, Reference [4] is shortly outlining how we can identify the person portrayed in a newly discovered artifact by comparing it with the images of books. In particular, she is telling that the collections of coins are quite useful. Accordingly, in [2] it is told that many scholars noted the lack of resemblances to Caesar's likenesses issued on coins. Moreover, there is also a lack of resemblance with the "Tusculum bust" of Caesar, depicting Julius Caesar in his lifetime. The possibility that the Arles bust was created during the 3rd century CE is also reported in [2].
The "Tusculum bust" is considered a realistic portrait of Julius Caesar. It was unearthed in 1825, in Tusculum and it is dated 50-40 BCE. It is permanently at the Museo d'Antichità, collezione del Castello Reale di Agliè, in Torino. In occasion of the exhibition, this bust is among the items from Turin, presented at the Louvre side by side with the Arles statue.
The Figure 1 is showing on the left, an image of the Arles bust and on the right the image of the Tusculum bust, from [2]. In fact, to use them, I prepared two grey tone images of the same size. I used the same black background too. For the Tusculum bust, GIMP smoothing and wavelets were quite useful, as I have discussed in some previous papers [5-9]. Here again, image processing is applied for multidisciplinary studies, as I made in Ref.7, where I used it to compare a self-portrait of Leonardo da Vinci with his portrait made by Raphael in the "School of Athens", where Leonardo appears as philosopher Plato. After processing the images of the two busts, we can have Figure 1. In this case, this image is not enough to conclude whether the two busts are portraits of the same man. To me, their features seem different.
The Tusculum bust is not the only Caesar's portrait we can use. There is another well-know statue, the Farnese bust, which is at the Archaeological Museum in Naples. It was found in Rome, in an area where, in medieval times, it was believed that the Basilica Ulpia was located. The bust portrays Julius Caesar with regular features, curly hair brushed towards his forehead, and two deep horizontal wrinkles. In Figure 2, I arranged three images: one of the Arles bust on the left, one of the Farnese bust on the right, and in the middle the two superimposed. In this case, it seems that the two men have the same features.

After this image processing, where I used a two-dimensional method, it seems that there is a certain agreement between Arles and Farnese busts. Therefore, I suggest that some further investigations and comparison of these three objects (Arles, Turin and Naples busts) could be interesting. A 3D scanning and rendering could give more details and parameters for comparison. Of course, this method is suitable for all the Caesar's busts available in museums and collections. After such processing, we could have some data and measures to create a virtual portrait of Julius Caesar, having some statistical meaning.

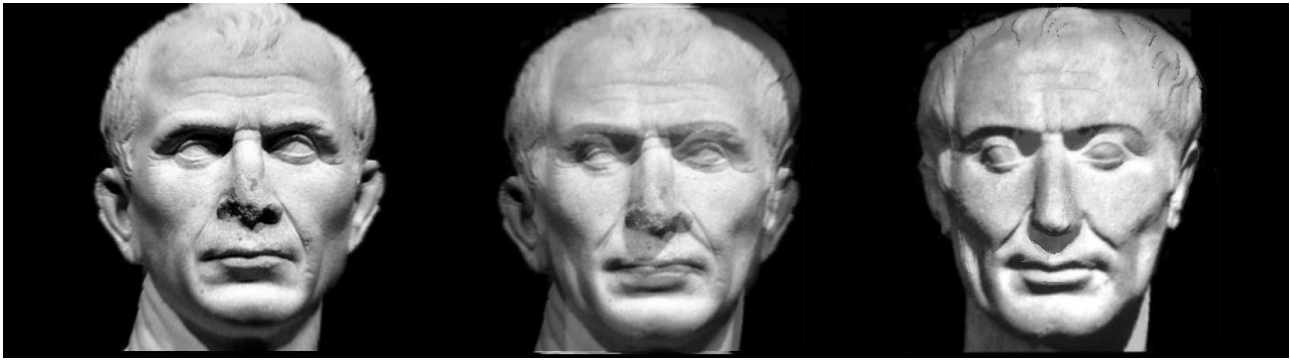

**Fig1: The Arles bust on the left and the Tusculum bust on the right. In the middle, the two images superposed (courtesy: Wikipedia).**

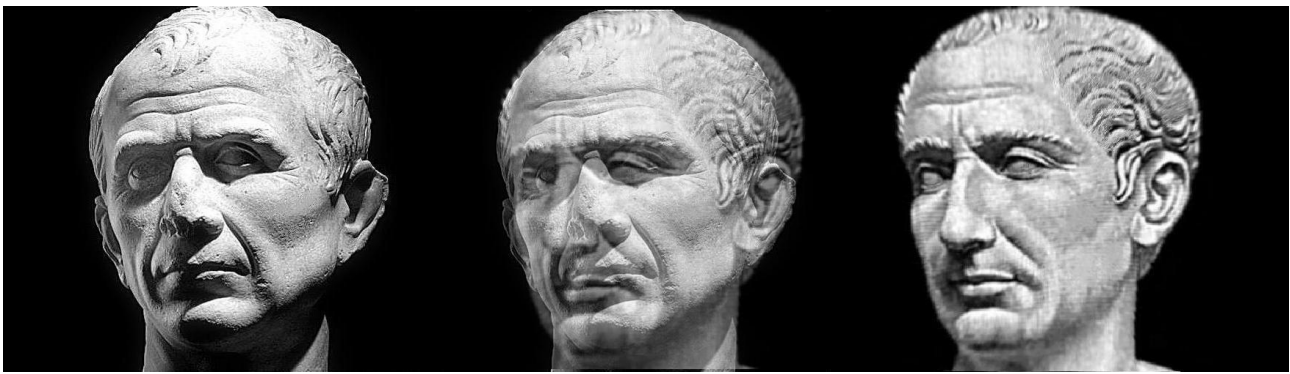

**Fig.2: The Arles bust on the left and the Farnese bust on the right. In the middle, the two images superposed (Courtesy, [1] and [11]).**